# Theoretically Accurate Regularization Technique for Matrix Factorization based Recommender Systems

Hao Wang
Ratidar.com
haow85@live.com

Regularization is a popular technique to solve the overfitting problem of machine learning algorithms. Most regularization technique relies on parameter selection of the regularization coefficient. Plug-in method and cross-validation approach are two most common parameter selection approaches for regression methods such as Ridge Regression, Lasso Regression and Kernel Regression. Matrix factorization based recommendation system also has heavy reliance on the regularization technique. Most people select a single scalar value to regularize the user feature vector and item feature vector independently or collectively. In this paper, we prove that such approach of selecting regularization coefficient is invalid, and we provide a theoretically accurate method that outperforms the most widely used approach in both accuracy and fairness metrics.

CCS CONCEPTS • Computing Methodologies • Machine Learning • Machine Learning Algorithms

**Additional Keywords and Phrases:** regularization, matrix factorization, recommender system, overfitting

## 1. INTRODUCTION

Overfitting is a commonly encountered problem in machine learning. When the number of parameters of the model is too large, the regression models usually captures too much noise and fail to generalize to new dataset. To solve the overfitting problem, there are many proposed algorithms in the machine learning field. For deep learning paradigms, drop-out is a popular technique that aims to reduce the complexity of the model structure. In the context setting of regressions, plug-in method and cross-validation approach are two classic techniques to compute the optimal regularization parameters.

Recommender system is a machine learning application field that is worth billions of money. There have been a tremendous amount of research publications in the field with innovative techniques such as deep learning models and hybrid methods. In recent years, ensemble tree models have also emerged as a major methodology for recommender system.

Regularization exists here and there in recommender system algorithms. For example, in the framework of matrix factorization, regularization has been used to reduce the norm of user feature vectors and item feature vectors. Common practice of selecting the regularization parameters have largely been ad-hoc, with a few notable exception. However, most people select a single constant regularization parameter to penalize the user feature vector and item feature vector together, or separate constants to penalize user feature vectors and item feature vectors by scalar-vector products.

In this paper, we prove that common practice of matrix factorization regularization has intrinsic flaws that cannot be resolved theoretically. We propose to regularize the user feature vector and item feature vector with separate regularization vectors. We use dot products of regularization vectors and feature vectors as the regularization term and we show how to achieve the optimal regularization parameters. In the experiment section, we demonstrate that our approach outperforms the common practice in both accuracy metric and fairness metric.

## 2 RELATED WORK

Recommender system is everywhere in modern day internet industry. Amazon, YouTube, Alibaba, Toutiao, TikTok all rely on recommendation products to generate values and cut short the marketing budgets. Matrix factorization is a mature and successful recommender system technology that has decade long history and mountainous applications in internet companies. Algorithms such as Alternating Least Squares [1] and SVDFeature [2], MatRec [3], Zipf Matrix Factorization [4], MatMat [5] and ZeroMat [6] are all variants of matrix factorization techniques.

Matrix factorization algorithms can be applied for context-aware recommendation [7], cold-start initialization [6], debiasing [4], in addition to its old mission of reducing the sum of squared error of the user rating approximation. It is a versatile algorithm that is worth of serious research even today. In this paper, we investigate into the regularization problem of matrix factorization to provide a superior and solid framework that is competitive on both MAE and fairness metrics.

Regularization is a popular technique not only used in matrix factorization but also regression [8][9] and ensemble tree models. Optimal regularization parameter selection problem has largely been solved in regression algorithms and incorporated into famous commercial softwares such as Stata. However, regularization problem outside of the regression scope has mostly been overlooked.

## 3 MATRIX FACTORIZATION REGULARIZATION

Matrix factorization with regularization is a common practice in commercial context settings. The classic matrix factorization with regularization is formulated as below :

$$L = \sum_{i=1}^{M}\sum_{j=1}^{N}\left(R_{i,j} - u_i^T \cdot v_j\right)^2 + \beta\left(\sum_{i=1}^{M}||u_i|| + \sum_{j=1}^{N}||v_j||\right)$$

We know that optimal regularization parameter for regression model exists and can be solved using plug-in method or cross-validation approach. We solve the optimal parameter value of β by taking partial derivative of β with respective to L, we obtain the following formula :

$$\frac{\partial L}{\partial u_i} = \sum_{j=1}^{N} 2(R_{i,j} - u_i^T \cdot v_j)v_j + \frac{\beta}{||u_i||}u_i$$

By setting this partial derivative to 0, we wish to find the minimizing value of $u_i$. But what is interesting is the the value of β here. We can actually solve for the value of β by letting the partial derivative to be 0 . We can compute β with M equations since there are M partial derivatives with respect to the user feature vectors :

$$\beta \begin{bmatrix} u_1^T \\ \ldots \\ u_M^T \end{bmatrix} = \begin{bmatrix} \sum_{j=1}^{M} 2||u_1||(R_{i,j} - u_1^T \cdot v_j)v_j \\ \ldots \\ \sum_{j=1}^{M} 2||u_M||(R_{i,j} - u_M^T \cdot v_j)v_j \end{bmatrix}$$

It is pretty obvious the equation system is not solvable for constant value of β in general. To see this point more clearly, we can rewrite the partial derivative equation as follows :

$$\frac{\beta}{||u_i||}u_i^T \cdot u_i + \sum_{j=1}^{N} 2(R_{i,j} - u_i^T \cdot v_j)u_i^T \cdot v_j = 0$$

Solving for the value of β, we obtain :

$$\beta = \frac{1}{||u_i||}\sum_{j=1}^{N} 2(R_{i,j} - u_i^T \cdot v_j)u_i^T \cdot v_j$$

It is obvious there is no single nonzero constant β for all different values of $u_i$ , otherwise we would have the following conclusion: Every single user's matrix factorization prediction value list for all his items weighted by approximation residuals is the norm of his user feature vector scaled by the same constant factor. This just sounds too idealistic and simplistic to us for us to believe this is true in real world commercial environments. Shouldn't the weighted sum have something to do with the item feature vector as well, rather than just the norm of the user feature vector scaled by a constant ? Why β is a constant for all different values of user feature vectors ?

We have shown that a single regularization parameter to penalize the norm of both user feature vector and item feature vector is an ill-posed problem. Penalizing the user feature vectors and item feature vectors using 2 constants does not change our conclusion, because we illustrated our idea using only partial derivative of user feature vector penalized by a constant above, the logic can be applied in the 2 regularization constant settings without any modification to our writings above.

Now we consider the following regularization approach and proves that it is also an ill-posed problem:

$$L = \sum_{i=1}^{M} \sum_{j=1}^{N} (R_{i,j} - u_i^T \cdot v_j)^2 + \left( \sum_{i=1}^{M} \beta_i ||u_i|| + \sum_{j=1}^{N} \gamma_j ||v_j|| \right)$$

We have the following equation by taking partial derivative with respect to the user feature vector:

$$\beta_i = \frac{1}{||u_i||} \sum_{j=1}^{N} 2(R_{i,j} - u_i^T \cdot v_j) u_i^T \cdot v_j$$

Once again, we are very reluctant to believe the validity of this formula because the sum of estimated user ratings of a single user weighted by approximation error has something to do only with the user feature vector nomr and has nothing to do with the item feature vector. The formula doesn't look right to us.

In this paper, we propose a new regularization framework as follows:

$$L = \sum_{i=1}^{M} \sum_{j=1}^{N} (R_{i,j} - u_i^T \cdot v_j)^2 + \left( \sum_{i=1}^{M} |\beta_i \cdot u_i| + \sum_{j=1}^{N} |\gamma_j \cdot v_j| \right)$$

, where $\beta_i$ and $\gamma_j$ are regularization vectors.

Using Stochastic Gradient Decesnt to solve for optimal values of parameters including the regularization vectors, we obtain the following formulas:

$$\frac{\partial L}{\partial u_i} = \sum_{j=1}^{N} 2(u_i^T \cdot v_j - R_{i,j}) v_j + sign(u_i^T \cdot \beta_i) \beta_i$$

$$\frac{\partial L}{\partial v_j} = \sum_{j=1}^{N} 2(v_j^T \cdot u_i - R_{i,j}) u_i + sign(v_j^T \cdot \gamma_j) \gamma_j$$

$$\frac{\partial L}{\partial \beta_i} = sign(u_i^T \cdot \beta_i) u_i$$

$$\frac{\partial L}{\partial \gamma_j} = sign(u_i^T \cdot v_j) v_j$$

If we solve for the vector $\beta_i$ using the partial derivative of the user feature vector as before, we would obtain the following formula :

$$||\beta_i||^2 = \frac{2(R_{i,j} - u_i^T \cdot v_j)\beta_i^T \cdot v_j}{\text{sign}(u_i^T \cdot \beta_i)}$$

The complexity of the formula makes us more willing to believe the validity of our regularization framework. We test our new framework of regularization on the LDOS-CoMoDa dataset [10] and the MovieLens dataset [11] to compare against the constant regularization framework. We show that our approach yields superior performance on both dataset and both the accuracy and fairness metrics.

## 4 EXPERIMENT

LDOS-CoMoDa data-set contains 121 users and 1232 movies with contextual information. We do not take advantage of the contextual information and just use the user rating matrix values for testing. We compare our new regularization framework (green color) against the constant regularization coefficient framework (blue color). We do grid search on the gradient descent learning steps and constant regularization coefficients and obtain the following results in MAE and Degree of Matthew Effect [4]:

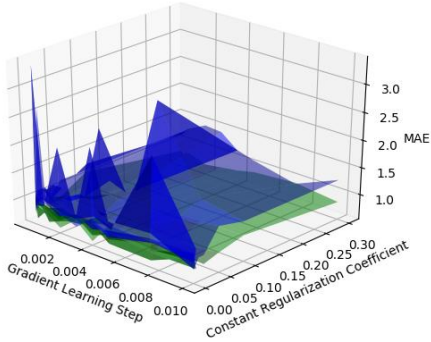
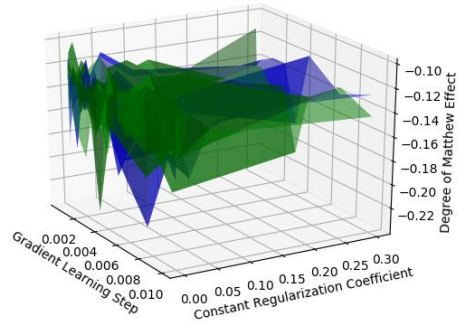

Fig. 1 Comparison between regularization frameworks on

Fig. 2 Comparison between regularization frameworks on

From the experiment results, we notice our new framework produces a much smoother surface of MAE scores and the MAE scores are much better than the constant parameter choice. The best MAE goes to 0.68, which is quite competitive compared with most recommender system algorithms. As for Degree of Matthew Effect, our algorithm is competitive as well, as illustrated in Fig. 2.

The MovieLens small dataset contains 610 users and 9724 movies. We conduct our experiments and illustrate the results below (Fig.3 and Fig. 4):

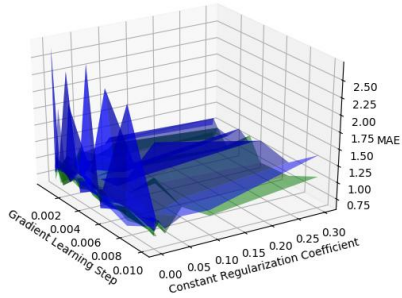

Fig. 3 Comparison between regularization frameworks on

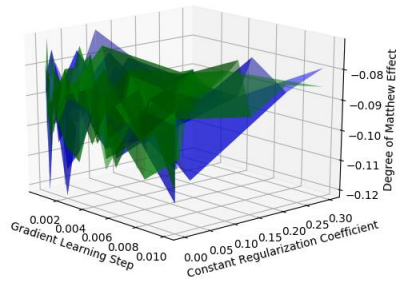

Fig. 4 Comparison between regularization frameworks on

Our observation from MovieLens small dataset is the same as LDOS-CoMoDa dataset. Our new framework achieves a best MAE score of 0.62, which is much better than the old regularization framework. Our method is also superior by fairness metric Degree of Matthew Effect.

We test our algorithms on large scale dataset for performance analysis. We use MovieLens 20M data set that consists of 138493 users and 26744 items. Our framework produces an MAE of 0.77 while the best MAE of constant regularizer is 0.80. Our framework produces smoother surface and more robust result overall. When evaluated in fairness metrics, the two frameworks are comparable.

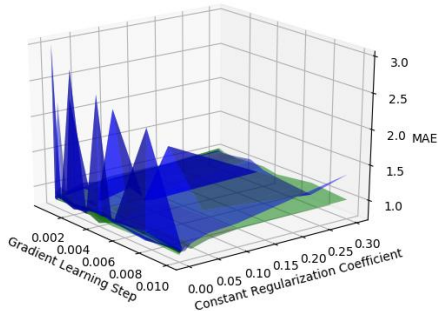

Fig. 5 Comparison between regularization frameworks on

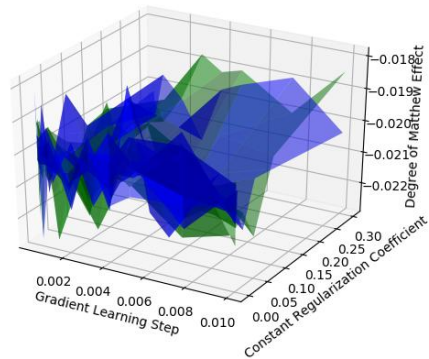

Fig. 6 Comparison between regularization frameworks on

## 5 DISCUSSION

We notice an interesting phenomenon for regularization frameworks: When we try to solve for regularization coefficients, whether they are constants or vectors, the assumption that regularization coefficient values are positive is contradictory to the minimization goal of the loss function. This interprets the reason why regularization solves the overfitting problem.

But on the other hand, this poses problems for our computation. We are optimizing in two opposite directions if we consider the regularization coefficients as variables to be optimized. If we apply the classic plug-in method to compute for the optimal values of regularization coefficients, the formulas look pretty weird, because the optimal values of user feature vector and item feature vector will lead to a zero valued regularization coefficient. This makes the parameter estimation of regularization coefficients using plug-in method a very difficult problem.

If we do not apply plug-in method, but consider regularization coefficients as variables, we would first need to answer the theoretical question whether the system of equations we try to solve is a well-posed problem, and what should we do if it is not. We notice there are literature on regularization of regression problems, but we need to pay more attention to other machine learning domain as well.

## 6 CONCLUSION

In this paper, we discuss the validity of popular regularization techniques of matrix factorization based recommender system. We prove by mathematical calculation that the common practice of regularization is not valid. We propose a different regularization technique that resorts to vector dot products to penalize the complexity of user feature and item feature vectors. We prove in the experiment section that our method outperforms the common practice in both MAE metric and Degree of Matthew Effect metric.

In future work, we would like to explore the possibility of finding the optimal form of regularization and selecting the best parameters. We believe our method could be generalized to other algorithms such as ensemble tree models.

## REFERENCES

[1] G. Takacs, D. Tikk. Alternating Least Squares for Personalized Ranking. The 6thACM Conference on Recommender Systems, 2012.
[2] T. Chen, W. Zhang, Q. Lu, K. Chen, Y. Yu. SVDFeature：A Toolkit for Feature-based Collaborative Filtering. The Journal of Machine Learning Research, 2012.
[3] H. Wang, B. Ruan. MatRec: Matrix Factorization for Highly Skewed Dataset. The 3rd International Conference on Big Data Technologies, 2020.
[4] H. Wang. Zipf Matrix Factorization: Matrix Factorization with Matthew Effect Reduction. The 4th International Conference on Artificial Intelligence and Big Data, 2021.
[5] H. Wang, MatMat: Matrix Factorization by Matrix Fitting. The 4$^{th}$ IEEE International Conference on Information Systems and Computer Aided Education, 2021. to appear.
[6] H. Wang, ZeroMat: Solving Cold-Start Problem of Recommender System with No Input Data. The 4$^{th}$ IEEE International Conference on Information Systems and Computer Aided Education, 2021. to appear.
[7] J. Zhao , W. Wang , Z. Zhang , et al. TrustTF: A tensor factorization model using user trust and implicit feedback for context-aware recommender systems, Knowledge-Based Systems, 2020
[8] D. Ruppert, S. Sheather, and M. Wand. An effective band-width selector for local least squares regression.Journal of the American Statistical Association, 1995.
[9] D. Ruppert and M. Wand. Multivariate locally weighted least squares regression. The Annals of Statistics, 1994

[10] T. Požrl. LDOS-CoMoDa dataset, 2012.
[11] T. Dong. MovieLens Dataset, 2017.